\documentclass[10pt,twocolumn,letterpaper]{article}

\usepackage[pagenumbers]{cvpr} 

\usepackage[dvipsnames]{xcolor}

\definecolor{cvprblue}{rgb}{0.21,0.49,0.74}
\usepackage[pagebackref,breaklinks,colorlinks,citecolor=cvprblue]{hyperref}
\usepackage{multirow}
\usepackage{colortbl}
\usepackage{pifont}
\usepackage{soul}
\usepackage{color, xcolor}
\definecolor{bgblue}{RGB}{231,239,252}
\definecolor{bgred}{RGB}{253,232,231}

\usepackage{xfrac}
\usepackage[accsupp]{axessibility}

\usepackage[ruled,linesnumbered]{algorithm2e}
\definecolor{commentcolor}{RGB}{110,154,155}   
\SetKwComment{Comment}{/* }{ */}
\SetKwInput{KwInput}{Input}
\SetKwInput{KwOutput}{Output}
\usepackage[capitalize]{cleveref}
\Crefname{algorithm}{Algorithm}{Algorithms}
\crefname{algorithm}{Alg.}{Algs.}

\def\paperID{1292} 
\def\confName{CVPR}
\def\confYear{2024}

\title{Efficient and Effective Weakly-Supervised Action Segmentation via Action-Transition-Aware Boundary Alignment}

\author{Angchi Xu$^{1}$,\quad Wei-Shi Zheng$^{1,2,3, *}$\\
$^1$School of Computer Science and Engineering, Sun Yat-sen University, China\\
$^2$Key Laboratory of Machine Intelligence and Advanced Computing, Ministry of Education, China\\
$^3$Guangdong Province Key Laboratory of Information Security Technology, Guangzhou, China\\
{\tt\small xuangch@mail2.sysu.edu.cn, wszheng@ieee.org}}

\begin{document}
{\onecolumn
\noindent \vspace{1cm}

\noindent \textbf{\huge{Efficient and Effective Weakly-Supervised Action \\Segmentation via Action-Transition-Aware \\Boundary Alignment}}

\vspace{2cm}

\noindent {\LARGE{Angchi Xu and Wei-Shi Zheng*}}
\\
\\
\large{*Corresponding author: Wei-Shi Zheng.}
\\
\\
\Large{Code: \url{https://github.com/iSEE-Laboratory/CVPR24_ATBA}}
\\
\\

\vspace{1cm}

\noindent {\Large{Accepted to IEEE/CVF Conference on Computer Vision and Pattern Recognition (CVPR) 2024}}

\vspace{1cm}

\noindent \large{For reference of this work, please cite:}

\vspace{1cm}
\noindent Angchi Xu and Wei-Shi Zheng. Efficient and Effective Weakly-Supervised Action Segmentation via Action-Transition-Aware Boundary Alignment. In \textit{Proceedings of the IEEE Conference on Computer Vision and Pattern Recognition}, 2024.

\vspace{1cm}

\noindent Bib:\\
\noindent @inproceedings\{xu2024efficient,\\
\ \ \  title     = \{Efficient and Effective Weakly-Supervised Action Segmentation via Action-Transition-Aware Boundary Alignment\}, \\
 \ \ \   author    = \{Xu, Angchi and Zheng, Wei-Shi\},\\
\ \ \  booktitle   = \{Proceedings of the IEEE/CVF Conference on Computer Vision and Pattern Recognition\},\\
\ \ \  year      = \{2024\}\\
\}
\thispagestyle{empty}
}
\twocolumn

\setcounter{page}{1}

\maketitle
\def\thefootnote{*}\footnotetext{Corresponding author.}\def\thefootnote{\arabic{footnote}}
\begin{abstract}
Weakly-supervised action segmentation is a task of learning to partition a long video into several action segments, where training videos are only accompanied by transcripts (ordered list of actions). Most of existing methods need to infer pseudo segmentation for training by serial alignment between all frames and the transcript, which is time-consuming and hard to be parallelized while training. 
In this work, we aim to escape from this inefficient alignment with massive but redundant frames, and instead to directly localize a few action transitions for pseudo segmentation generation, where a transition refers to the change from an action segment to its next adjacent one in the transcript.
As the true transitions are submerged in noisy boundaries due to intra-segment visual variation, we propose a novel Action-Transition-Aware Boundary Alignment (ATBA) framework to efficiently and effectively filter out noisy boundaries and detect transitions. In addition, to boost the semantic learning in the case that noise is inevitably present in the pseudo segmentation, we also introduce video-level losses to utilize the trusted video-level supervision.
Extensive experiments show the effectiveness of our approach on both performance and training speed.\footnote{Code is available at \url{https://github.com/iSEE-Laboratory/CVPR24_ATBA}.}
\end{abstract}

\section{Introduction}
\label{sec:intro}

Action segmentation aims to partition a long untrimmed video into several segments and classify each segment into an action category~\cite{farha2019ms, souri2021fast, rahaman2022generalized, lu2022set, du2022fast, ghoddoosian2023weakly, yang2023lac, ghoddoosian2022weakly, shen2022semi, ding2022leveraging}. It is an important yet challenging task for instructional or procedural video understanding. Although fully-supervised action segmentation (FSAS) methods~\cite{farha2019ms, gao2021global2local, chen2020action, liu2023diffusion, bahrami2023much} have achieved great progress, they require frame-wise dense annotation, which is labor-intensive and time-consuming to collect. As a result, many works~\cite{li2019weakly, lu2021weakly, souri2021fast, chang2021learning, richard2018neuralnetwork, ding2018weakly, zhang2023hoi} explore the weakly-supervised action segmentation (WSAS) only requiring the \textit{transcript} annotation, which refers to the ordered list of actions occurring in the video without their start and end times. The transcripts are less costly to obtain and can be accessed directly from video narrations or other meta data~\cite{lu2021weakly, ding2018weakly, richard2017weakly, huang2016connectionist, bojanowski2014weakly, miech2019howto100m, zhukov2019cross}.

Most of previous WSAS methods have to infer the pseudo segmentation (pseudo frame-wise labels) for training via a sequence alignment process between the video and given transcript, such as Viterbi~\cite{richard2017weakly, richard2018neuralnetwork, lu2021weakly, li2019weakly, kuehne2017weakly, kuehne2018hybrid} or Dynamic Time Warping (DTW)~\cite{chang2019d3tw, chang2021learning}. These alignment algorithms are usually designed in a recursive form which needs to be performed serially frame-by-frame and hard to be parallelized, resulting in very slow training process. 

\begin{figure}
    \centering
    \includegraphics[width=\linewidth]{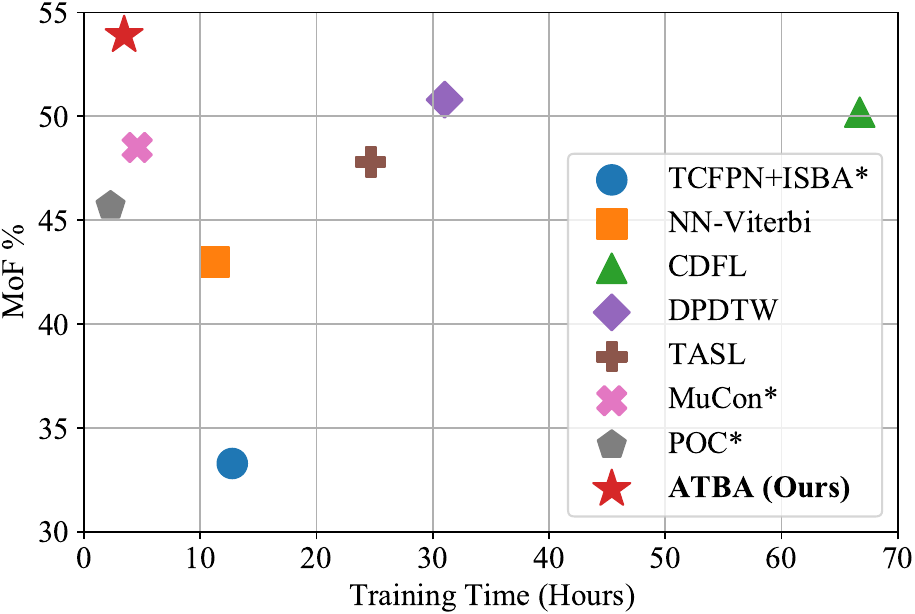}
    \caption{Comparison of performance and training time of WSAS methods on the Breakfast dataset. MoF-The main metric of the task, the higher the better. *-Alignment-free methods. Our ATBA achieves the best performance with a very short training time.}
    \label{fig:perf}
\end{figure}

\begin{figure}
    \centering
    \includegraphics[width=\linewidth]{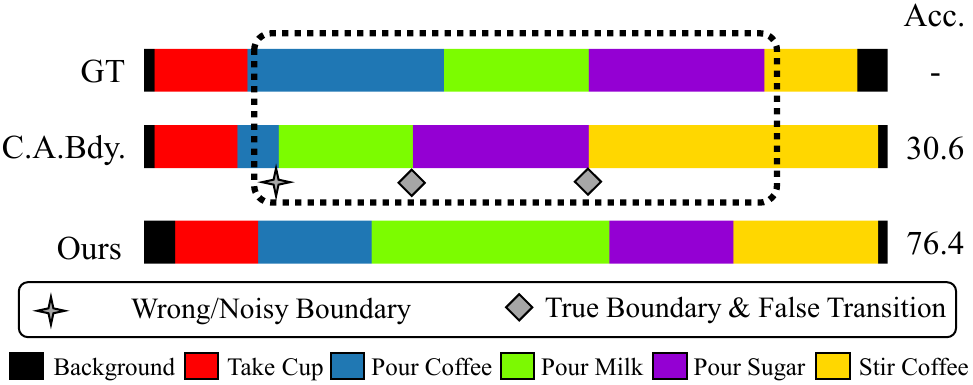}
    \caption{The necessity of proposed ATBA. The example is \textit{P54-webcam01-P54-coffee} in Breakfast dataset. GT-The ground-truth segmentation. C.A.Bdy.-Only class-agnostic boundary detection is applied (Exp.1 of \Cref{tab:atba}). Acc.-The accuracy of pseudo segmentation. In the video clip around the “star” point, the coffee pot undergoes a change from being picked up to tilted pouring within the segment ``\textit{Pour Coffee}", and this noisy visual change is incorrectly detected. In addition, although two boundaries are correctly detected by the ``C.A.Bdy." (diamonds), they correspond to incorrect transitions due to one false positive error (star), resulting in complete dislocation of segments within the dashed box. Best viewed in color.}
    \label{fig:intro_bdy}
\end{figure}

In this work, we argue that the frame-by-frame alignment is NOT necessary, since the pseudo segmentation is fundamentally determined by the locations of a \textit{small} number of \textit{action transitions} (\ie, the change from an action segment to its \textit{next adjacent} action segment in the transcript). Hence, the pseudo segmentation generation can be viewed as a transition detection problem, implying the way to more efficient designs. Intuitively, action transitions are often accompanied by significant visual changes, and there are already many approaches that can detect class-agnostic action \textit{boundaries} based on these changes \cite{kang2022uboco, du2022fast}. However, due to the intra-segment visual variation and sub-segments under finer granularity, there are numerous \textit{noisy} boundaries not corresponding to any transitions. Moreover, as the class-agnostic way cannot guarantee correct correspondence between the boundaries and transitions, even slight errors in the detection can result in severe deviation  (\cref{fig:intro_bdy}).

To overcome the above noisy boundary issue, we propose an efficient and effective framework for WSAS, termed \textit{\textbf{A}ction-\textbf{T}ransition-Aware \textbf{B}oundary \textbf{A}lignment} (ATBA), which directly detects the transitions for faster and effective pseudo segmentation generation. To tolerate the noisy boundaries, the ATBA generates \textit{more} class-agnostic boundaries than the number of transitions as candidates, and then determines a subset from candidates that \textit{optimally} matches all desired transitions via a drop-allowed alignment algorithm. Furthermore, to fortify the semantic learning under the inevitable noise in pseudo segmentation, we also introduce video-level losses to make use of the \textit{trusted} video-level supervision. Our ATBA is efficient, because the number of generated candidates will be proportional to the length of the transcript, and therefore the complexity of alignment is now \textit{independent} of the very long video length. Moreover, other computations required by ATBA, \ie, measuring \textit{how likely a frame is to be a boundary} and \textit{how likely a candidate corresponds to a desired transition}, are both built on a convolution-like algorithm inspired by \cite{kang2022uboco}, which can be parallelized on GPUs efficiently. 

For inference, we directly adopt the results from the trained frame-wise classifier, without the need for any alignment processing with retrieved or predicted transcript (the ground-truth transcript is not available during inference) like previous WSAS methods \cite{richard2018neuralnetwork, li2019weakly, chang2019d3tw, lu2021weakly, chang2021learning, souri2021fast}, which also improves the inference efficiency.

In summary, our contributions are as follows. (1) We propose to directly localize the action transitions for efficient pseudo segmentation generation during training, without the need of time-consuming frame-by-frame alignment. (2) For robustness to noisy boundaries, we propose a novel ATBA framework to effectively determine boundaries corresponding to each transition. Video-level losses are also introduced to regularize the semantic learning involving the unavoidable noise in the pseudo segmentation. Experiments are conducted on three popular datasets to evaluate our approach: Breakfast~\cite{kuehne2014language}, Hollywood Extended~\cite{bojanowski2014weakly} and CrossTask~\cite{zhukov2019cross}. Our ATBA achieves state-of-the-art or comparable results with one of the fastest training speed (\cref{fig:perf}), demonstrating the effectiveness of ours.

\section{Related Work}
\label{sec:related}
Weakly-supervised action segmentation methods learn to partition a video into several action segments from  training videos only annotated by transcripts~\cite{bojanowski2014weakly, huang2016connectionist, richard2018neuralnetwork, ding2018weakly, li2019weakly, chang2019d3tw, richard2017weakly, souri2021fast, chang2021learning, lu2021weakly, kuehne2017weakly, souri2022fifa, kuehne2018hybrid, zhang2023hoi}. Despite different optimization objectives, most of them generate the pseudo segmentation for training by solving alignment objectives between two sequences (video and transcript) via Connectionist Temporal Classification (CTC)~\cite{huang2016connectionist}, Viterbi~\cite{richard2017weakly, richard2018neuralnetwork, li2019weakly, lu2021weakly, kuehne2017weakly, kuehne2018hybrid} or Dynamic Time Warping (DTW)~\cite{chang2019d3tw, chang2021learning}.

\begin{figure*}
    \centering
    \includegraphics[width=1.0\linewidth]{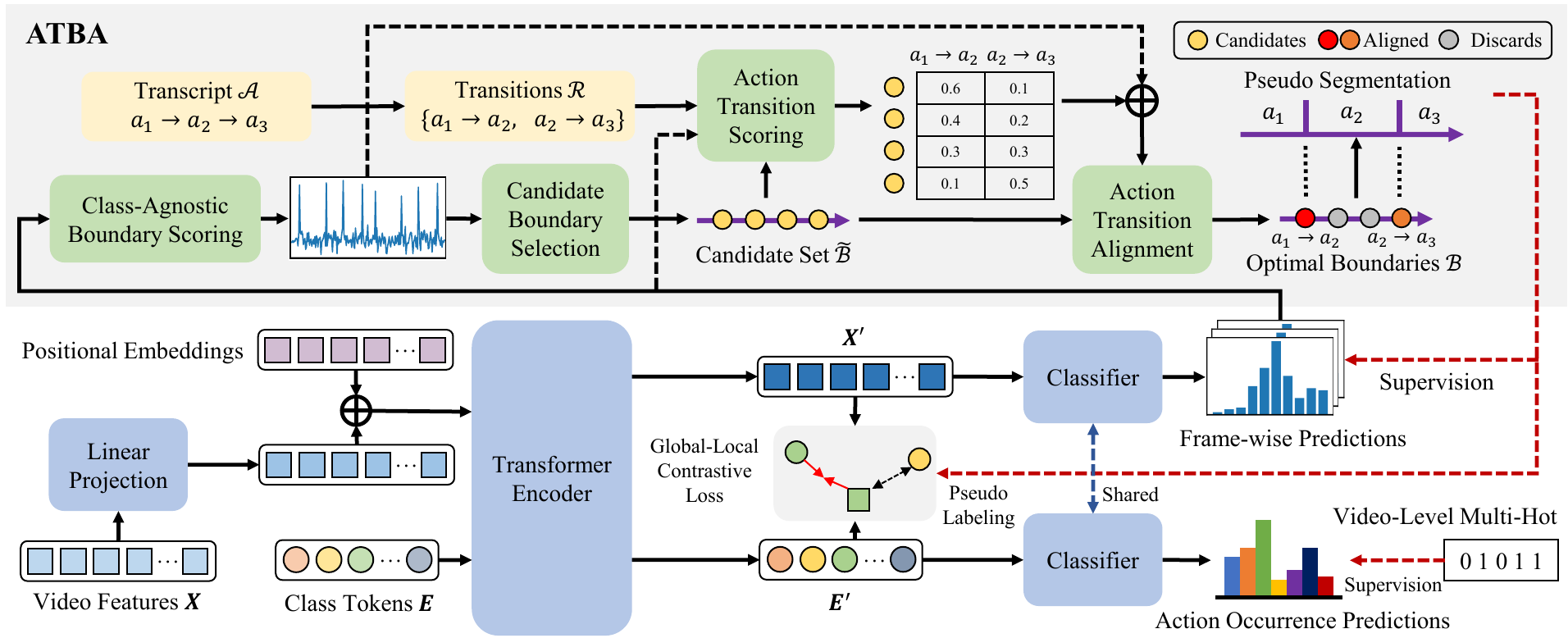}
    \caption{The overall framework. We propose an Action-Transition-Aware Boundary Alignment (ATBA) framework, which takes the class-agnostic boundary pattern and action transition pattern together into account to efficiently generate pseudo labels. The trusted video-level supervision is also utilized to further enhance the performance.}
    \label{fig:framework}
\end{figure*}

Specifically, \cite{huang2016connectionist} proposes an extended version of CTC to evaluate all valid alignments between videos and transcripts, which additionally takes the visual similarities of frames into account. Inspired by speech recognition, \cite{kuehne2017weakly, richard2017weakly, kuehne2018hybrid} all use the Hidden Markov Model (HMM) to model the relationship between videos and actions. \cite{ding2018weakly} always generates uniform segmentation but iteratively adjusts the boundaries by inserting repeating actions into the transcript. \cite{richard2018neuralnetwork} proposes an alignment objective based on explicit context and length models, which can be solved by Viterbi, and the solved optimal alignment would serve as pseudo labels to train the frame-wise classifier. \cite{li2019weakly} and \cite{lu2021weakly} both focus on novel learning objectives, but still require the optimal pseudo segmentation produced by Viterbi. \cite{chang2019d3tw, chang2021learning} both learn from the contrast of aligning the video to the ground-truth transcript and negative transcripts, where the alignment is performed by DTW. Whether using CTC, Viterbi or DTW, the above approaches except~\cite{ding2018weakly} require frame-by-frame serial calculation, which are inefficient.

Recently, some efficient methods are proposed with alignment-free design. \cite{souri2021fast} learns from the mutual consistency between two forms of a segmentation (\ie, frame-wise classification and category/length pairs). \cite{lu2022set} proposes a loss to enforce the output order of any two actions to be consistent with the transcript.\footnote{POC~\cite{lu2022set} is a set-supervised method but can be extended to transcript supervision naturally, so we cite its corresponding results for comparison.} In this work, we also propose an efficient framework with different technical roadmap, and our performance is better at the same level of training speed.

\section{Approach}

\subsection{Problem Statement}
Action segmentation is a task of partitioning a video into several temporal segments with action labels, which is equivalent to predicting the action categories of \textit{each} frame.
Formally, given a sequence of $T$ $d$-dimensional frame-wise features $\boldsymbol{X}=[\boldsymbol{x}_1, ..., \boldsymbol{x}_T] \in \mathbb{R}^{T\times d}$ for a video with $T$ frames, the goal is to predict a sequence of actions $\hat{\mathcal{Y}}=[\hat{y}_1, ..., \hat{y}_T]$, where $\hat{y}_t \in \mathcal{C}$ and $\mathcal{C}=\{1,2,...,|\mathcal{C}|\}$ is the set of action categories across the dataset (including the \textit{background}). Under the setting of WSAS, the frame-wise ground-truth $\mathcal{Y}=[y_1, ..., y_T]$ is NOT available during training. Instead, the ordered list of actions called transcript $\mathcal{A}=[a_1, ..., a_M]$ is provided (including the background segments), where $a_m \in \mathcal{C}$ and $M$ is the total number of action segments in the video. The action transitions of $\mathcal{A}$ are naturally formulated as $\mathcal{R}=\{(a_r, a_{r+1})\}_{r=1}^{M-1}$.

\subsection{Overview}
Our proposed framework is illustrated in \cref{fig:framework}. At first, the input sequence $\boldsymbol{X}$ is further encoded by a temporal network to generate more task-relevant representations $\boldsymbol{X}'\in \mathbb{R}^{T\times d'}$ (\cref{sec:network}), then a classifier shared along the temporal axis followed by a category softmax activation will predict the frame-wise class probabilities $\boldsymbol{P}=[\boldsymbol{p}_1,...,\boldsymbol{p}_T] \in \mathbb{R}^{T\times|\mathcal{C}|}$ from $\boldsymbol{X}'$. After that, the Action-Transition-Aware Boundary Alignment (ATBA) module takes $\boldsymbol{P}$ and the transcript $\mathcal{A}$ as input to infer the pseudo frame-wise labels $\widetilde{\mathcal{Y}}=[\widetilde{y}_1,...,\widetilde{y}_T]$, where $\widetilde{y}_t\in \mathcal{C}$  (\cref{sec:atba}). Finally, $\widetilde{\mathcal{Y}}$ is used back to supervise $\boldsymbol{P}$ by a standard cross entropy:
\begin{equation}
\mathcal{L}_{\text{cls}}=-\dfrac{1}{T}\sum_{t=1}^T\sum_{c=1}^{|\mathcal{C}|}\mathbb{I}(\widetilde{y}_t=c)\log\boldsymbol{P}_{t,c},
\label{eq:cls}
\end{equation}
where $\mathbb{I}(\cdot)$ is the indicator function which returns 1 if the condition is satisfied and 0 otherwise. Note that the pseudo frame-wise labels are inferred at each iteration for a batch of data and used immediately for current training at the same iteration. 
The additional video-level losses are stated in \cref{sec:loss}, and the training and inference processes are described in \cref{sec:inf}.

\subsection{Temporal Network}
\label{sec:network}
We employ a slightly modified pre-norm \cite{xiong2020layer} Transformer \cite{vaswani2017attention} encoder with learnable positional embeddings as the temporal network for feature learning. Following \cite{yi2021asformer, du2022efficient}, the vanilla \textit{full} self-attention is replaced with a pyramid hierarchical \textit{local} attention to better adapt to the action segmentation task (see supplemental material for more details). 

\subsection{Action-Transition-Aware Boundary Alignment}
\label{sec:atba}
The ATBA is the core component of our framework. It generates the pseudo frame-wise labels $\widetilde{\mathcal{Y}}$ by inferring the boundaries corresponding to $M-1$ action transitions from the predicted class probabilities $\boldsymbol{P}$. Once they are found, $\widetilde{\mathcal{Y}}$ can be naturally generated by assigning the action labels of $\mathcal{A}$ one-by-one into the intervals between these boundaries.

Briefly, the ATBA first generates a set of candidate boundaries via a class-agnostic way, and then finds $M-1$ points from this set that optimally match all transitions of $\mathcal{A}$ via a dynamic programming (DP) algorithm. We describe the details in the following.

\begin{figure}
    \centering
    \includegraphics[width=0.8\linewidth]{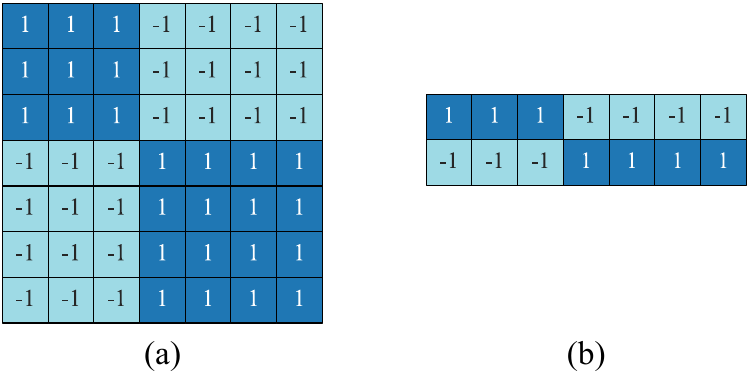}
    \caption{(a) A 7$\times$7 template for class-agnostic boundary scoring. (b) A 2$\times$7 template for action transition scoring.}
    \label{fig:kernel}
\end{figure}

\noindent\textbf{- Class-Agnostic Boundary Scoring.} Firstly, the ATBA calculates the class-agnostic boundary scores $\mathcal{V}^\text{b}=[v^\text{b}_1,...,v^\text{b}_T]$ for \textit{each} timestamp. We employ a pattern-matching-based scoring method proposed by a generic event boundary detection approach termed UBoCo~\cite{kang2022uboco}, which considers the pattern of each frame's neighborhood. Specifically, for timestamp $t$, a pairwise similarity matrix $\boldsymbol{\Gamma}^{(t)} \in \mathbb{R}^{w^\text{b}\times w^\text{b}}$ is calculated within a local window with size $w^\text{b}$ centered at $t$, whose $(i,j)$-entry represents the class-agnostic similarity between $i$-th and $j$-th frames inside the window, with values ranging from -1 to 1 (see the computational details in supplementary material). Clearly, if $t$ is a boundary, the frame feature should change dramatically at $t$ and keep stable elsewhere, so its $\boldsymbol{\Gamma}^{(t)}$ should show the pattern of that the values in the upper-left and lower-right areas are close to 1 and otherwise close to -1. Hence a template $\boldsymbol{\Omega}^\text{b} \in \mathbb{R}^{w^\text{b}\times w^\text{b}}$ like \cref{fig:kernel}(a) is designed to capture this pattern and output the class-agnostic boundary score for each~$t$ by a correlation operation:
\begin{equation}
    v^\text{b}_t = \dfrac{1}{w^\text{b}\times w^\text{b}}\sum_{i=1}^{w^\text{b}}\sum_{j=1}^{w^\text{b}}\boldsymbol{\Omega}^\text{b}_{i,j} \boldsymbol{\Gamma}^{(t)}_{i,j}.
\end{equation}

\noindent\textbf{- Candidate Boundary Selection.} After calculating $\mathcal{V}^\text{b}$, we select a set of $K$ candidate boundaries $\widetilde{\mathcal{B}}=\{b_k\}_{k=1}^K$, where $1 < b_1 < ... < b_K \leq T$ and $K > M-1$. The selection is performed by a simple greedy strategy with non maximum suppression inspired by \cite{du2022fast}, \ie, each time we select one timestamp with current highest score $v^\text{b}_t$, and invalidate its neighborhood to avoid selecting multiple timestamps corresponding to one same boundary, until the number of selected timestamps reach an upper bound or all remaining timestamps are invalid. The radius of the invalid interval is set adaptively to $\frac{\mu T}{M}$, where $\mu \in [0,1]$ is a hyper-parameter, and in practice, the upper bound for $K$ is set to $\lambda(M-1)$, where $\lambda \in \mathbb{N}_+$ is also a hyper-parameter.

\noindent\textbf{- Action Transition Scoring.} As mentioned in \cref{sec:intro}, the class-agnostic scores $\mathcal{V}^\text{b}$ are not enough to detect action transitions.
To this end, we then calculate an action transition score matrix $\boldsymbol{V}^\text{a} \in \mathbb{R}^{K\times (M-1)}$, where $\boldsymbol{V}^\text{a}_{k, r}$ measures the possibility that the $k$-th candidate boundary corresponds to the $r$-th transition, \ie, separates the $r$-th and $(r+1)$-th action segments.
This matrix is also calculated via pattern matching. Clearly, if candidate $b_k$ corresponds to the $r$-th transition, the classifier's activation for class $a_r$ should drop sharply after $b_k$, while rise for class $a_{r+1}$.
Hence, a template $\boldsymbol{\Omega}^\text{a} \in \mathbb{R}^{2\times w^\text{a}}$ with temporal size $w^\text{a}$ like \cref{fig:kernel}(b) is employed to detect this pattern around $b_k$:
\begin{equation}
\begin{gathered}
    \boldsymbol{V}^\text{a}_{k, r} = \dfrac{1}{2w^\text{a}}\sum_{i=1}^{2}\sum_{j=1}^{w^\text{a}}\boldsymbol{\Omega}^\text{a}_{i,j}\boldsymbol{P}_{\text{ind}^\text{a}(b_k, j),\ a_{r+i-1}},\\
    \text{ind}^\text{a}(b_k,j)=b_k-\lfloor\dfrac{w^\text{a}}{2}\rfloor+j-1,
\end{gathered}
\end{equation}
where $\text{ind}^\text{a}(b_k,j)$ is the index transform from the index $j$ of the local window centered at $b_k$ to the global timestamp index. Finally, the class-agnostic scores $\mathcal{V}^\text{b}$ are added back to $\boldsymbol{V}^\text{a}$ to produce the final score matrix $\boldsymbol{V}\in\mathbb{R}^{K\times (M-1)}$:
\begin{equation}
    \label{eq:add}
    \boldsymbol{V}_{k, r} = \boldsymbol{V}^\text{a}_{k, r} + v^\text{b}_{b_k}.
\end{equation}
The above equation means that the class-agnostic boundary score of the $k$-th boundary $v^\text{b}_{b_k}$ is added to \textit{all} transition scores corresponding to the $k$-th boundary $\boldsymbol{V}^\text{a}_{k, r}, \forall r$.

\begin{figure}
    \centering
    \includegraphics[width=0.8\linewidth]{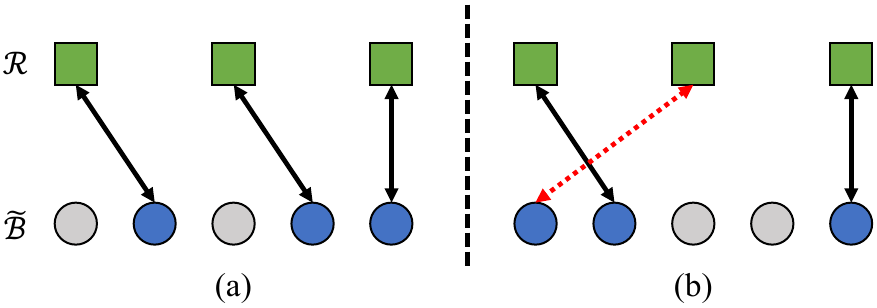}
    \caption{Illustration of the alignment between action transitions $\mathcal{R}$ and candidate boundaries $\widetilde{\mathcal{B}}$. Blue circles are aligned and gray ones are dropped. (a) A valid alignment. (b) An invalid alignment. The red dashed arrow violates the ordering consistency. Best viewed in color.}
    \label{fig:align}
\end{figure}

\noindent\textbf{- Action Transition Alignment.} The final step is to find exact $M-1$ optimal boundaries from the candidates $\widetilde{\mathcal{B}}$ based on the score matrix $\boldsymbol{V}$. It is equivalent to seek an one-to-one alignment with lowest cost between the transitions $\mathcal{R}$ and candidates $\widetilde{\mathcal{B}}$ while requiring $K-M+1$ candidates to be dropped (\cref{fig:align}). Formally, if we denote the optimal aligned boundary set as $\mathcal{B}=\{b_{k_r}\}_{r=1}^{M-1}$, where $b_{k_r}$ corresponds to the $r$-th transition and $1\leq k_1 < ... < k_{M-1} \leq K$. It should satisfy:
\begin{equation}
    \mathcal{B}=\arg\min_{\mathcal{B}'}\psi(\mathcal{B}'),\quad
    \psi(\mathcal{B}') = -\sum_{r=1}^{M-1}\boldsymbol{V}_{k_r',r},
    \label{eq:align}
\end{equation}
where $\psi(\mathcal{B}')$ is the cost function of an alignment and $\mathcal{B}'$ is any feasible aligned boundary set such as $\mathcal{B}$.

\begin{figure}
    \centering
    \includegraphics[width=\linewidth]{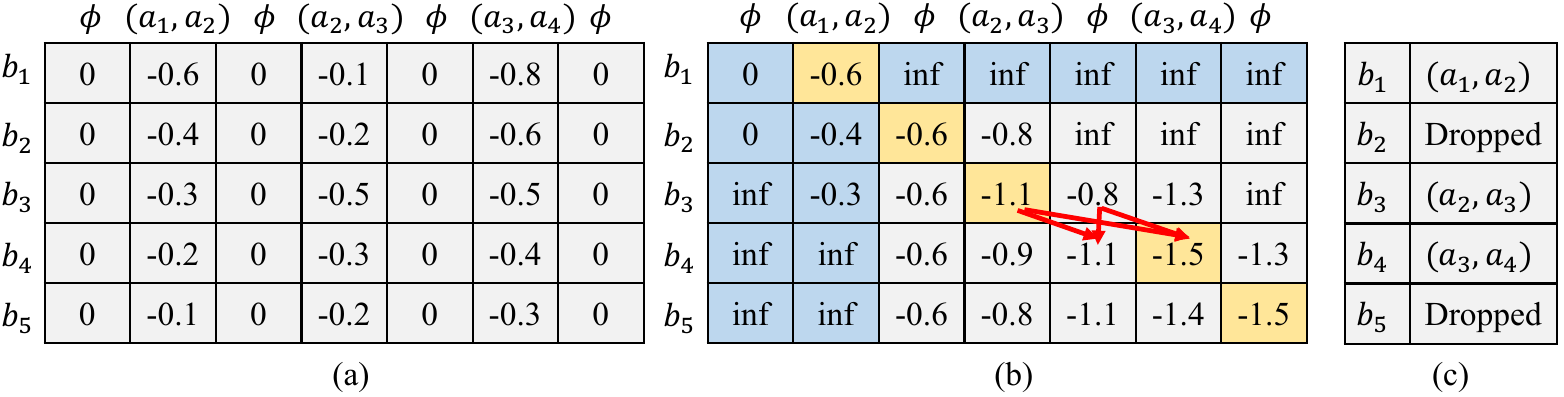}
    \caption{An example of action transition alignment between $\widetilde{\mathcal{B}}$ of length 5 and $\mathcal{R}$ of length 3. (a) The cost matrix $\boldsymbol{\Delta}$. (b) The cumulative cost matrix $\boldsymbol{D}$. Blue areas are directly initialized. Yellow areas correspond to optimal alignment. Red arrows are allowed path directions. (c) The optimal alignment. Best viewed in color.}
    \label{fig:dtw}
\end{figure}

To solve \cref{eq:align}, we propose a DP algorithm, of which an example is shown in \cref{fig:dtw}. To allow some candidates in $\widetilde{\mathcal{B}}$ to be dropped, inspired by~\cite{shen2021learning}, we first expand the transition sequence $\mathcal{R}$ by inserting the \textit{empty} symbol $\phi$ interleaved with transition symbols, obtaining $\mathcal{R}'=[\phi, (a_1,a_2),\phi,(a_2,a_3),\phi,...,\phi,(a_{M-1}, a_M),\phi]$. Now the candidate boundary matched to $\phi$ will be discarded. Then, a cost matrix $\boldsymbol{\Delta}\in \mathbb{R}^{K\times (2(M-1)+1)}$ is constructed (\cref{fig:dtw}(a)), whose $(k, r')$-th entry is the cost of aligning the $b_k$ with the $r'$-th symbol in $\mathcal{R}'$. For even number of $r'$ (transition symbols), it is clear that $\boldsymbol{\Delta}_{k,r'}=-\boldsymbol{V}_{k,r'/2}$ from \cref{eq:align}. Odd $r'$ means dropping the candidate, and the cost is set to 0.

A valid alignment is represented by a path from the top-left to bottom-right of $\boldsymbol{\Delta}$ with constraint on directions. Specifically, the allowed start positions are $(1, 1)$ and $(1, 2)$, respectively corresponding to two cases where $b_1$ is dropped and $b_1$ is matched with the first transition. Similarly, there are two allowed end positions: $(K,2(M-1))$ and $(K,2(M-1)+1)$. Then for the position $(k, r')$ on the path, if $r'$ is odd, its previous position is either $(k-1, r')$ or $(k-1, r'-1)$. It means that if $b_k$ is dropped, the $b_{k-1}$ is also dropped or matched with the previous transition. On the other hand, the previous position is either $(k-1,r'-1)$ or $(k-1, r'-2)$ if $r'$ is even, with the similar meaning. These allowed directions are indicated by red arrows in \cref{fig:dtw}(b).

The DP algorithm generally finds the optimal solution by computing a cumulative cost matrix $\boldsymbol{D}$ with the same shape as $\boldsymbol{\Delta}$ (\cref{fig:dtw}(b)), where $\boldsymbol{D}_{k,r'}$ represents the \textit{minimum} cumulative cost of all valid paths \textit{ending} at $(k,r')$. It is computed by the following recursive equation, considering paths \textit{coming from} valid directions aforementioned:
\begin{equation}
    \boldsymbol{D}_{k,r'}=\boldsymbol{\Delta}_{k,r'}+\biggl\{
    \begin{aligned}
    \min(\boldsymbol{D}_{k-1,r'},\boldsymbol{D}_{k-1,r'-1}),\ &r' \text{is odd},\\
    \min(\boldsymbol{D}_{k-1,r'-1},\boldsymbol{D}_{k-1,r'-2}),\ &r' \text{is even}.
    \end{aligned}
\end{equation}
The above equation is not applicable to the first row and the first two columns of $\boldsymbol{D}$. They are directly initialized before the computation (blue areas of \cref{fig:dtw}(b)). We set the cost of invalid positions to $\infty$ (inf) and others are copied from $\boldsymbol{\Delta}$. 

The cumulative cost of the optimal alignment is $\min(\boldsymbol{D}_{K,2(M-1)}, \boldsymbol{D}_{K,2(M-1)+1})$, 
The complete alignment, \ie, the optimal boundary set $\mathcal{B}$ can be obtained by starting from the corresponding position of the above \texttt{min} operation and backtracking (\cref{fig:dtw}(b/c)). 
More details can be found in the supplementary material.

\noindent\textbf{- Complexity.} Clearly, the class-agnostic boundary scoring and action transition scoring can be implemented by the \texttt{unfold} operator and matrix operations, thus can be parallelized on GPUs. The action transition alignment must be performed serially with a time complexity of $O(KM)$ or $O(\lambda M^2)$. Compared with the previous alignment methods such as Viterbi ($O(T^2M)$~\cite{richard2018neuralnetwork, li2019weakly, lu2021weakly}) and DTW ($O(TM)$), it decreases from a function involving video length $T$ to one of \textit{only} transcript length $M$ ($M << T$).

\subsection{Video-Level Losses}
\label{sec:loss}

Inevitably, the inferred pseudo segmentation contains some degree of noise, which can be unexpectedly fit by the network. To improve the semantic robustness, we propose to jointly train a video-level multi-label classification task that predicts \textit{whether each action appears in the video}, which is \textit{precisely} indicated by the transcript. Specifically, inspired by the common practice of Transformer \cite{lanchantin2021general,xu2022multi, devlin2018bert, dosovitskiy2020image, jiang2021all}, the input sequence $\boldsymbol{X}$ is augmented by a set of $|\mathcal{C}|$ learnable class tokens which are fed into the same network and responsible for predicting action occurrence.
The outputs, denoted as $\{\boldsymbol{e}'_c\}_{c=1}^{|\mathcal{C}|}$, are then used to predict the action occurrence probability via $\xi_c = \sigma(\boldsymbol{w}_c^T\boldsymbol{e}'_c + \epsilon_c)$ for each category $c$, where $\boldsymbol{w}_c$ and $\epsilon_c$ are the weight vector and bias scalar for class $c$ in the classifier \textit{shared} with frame-wise classification. $\sigma(\cdot)$ is the sigmoid activation. These predictions are supervised by the binary cross entropy:
\begin{equation}
    \mathcal{L}_\text{vid}=-\dfrac{1}{|\mathcal{C}|}\sum_{c=1}^{|\mathcal{C}|}[y^\text{vid}_c\log\xi_c+(1-y^\text{vid}_c)\log(1-\xi_c)],
\end{equation}
where $y^\text{vid}_c \in \{0,1\}$ is the binary label indicating whether action $c$ appears in the video. Through the shared network and the interactions between tokens, the semantics learned from video-level training can benefit the main task.

Moreover, we treat the class tokens with global information as the \textit{prototypes} of each action and attempt to align the frame features to them for semantic consistency. Specifically, we first calculate the feature centroid of each category $c$ that appears in the video by averaging the output frame features with the corresponding pseudo labels:
\begin{equation}
    \overline{\boldsymbol{x}}_c = \dfrac{\sum_{t=1}^{T}\mathbb{I}(\widetilde{y}_t=c)\boldsymbol{x}'_t}{\sum_{t=1}^{T}\mathbb{I}(\widetilde{y}_t=c)},\ c\in\text{Set}(\mathcal{A}),
\end{equation}
where the $\text{Set}(\cdot)$ operator converts a list into an unordered set and removes duplicates. These centroids will move towards the corresponding $\boldsymbol{e}'_c$ under the guidance of a global-local contrastive loss in the form of InfoNCE~\cite{he2020momentum, oord2018representation}:
\begin{equation}
    \mathcal{L}_\text{glc}=-\dfrac{1}{|\text{Set}(\mathcal{A})|}\sum_{c\in\text{Set}(\mathcal{A})}\log\dfrac{\exp(\langle\overline{\boldsymbol{x}}_c\rangle^T\langle\boldsymbol{e}'_c\rangle / \tau)}{\sum_{c'=1}^{|\mathcal{C}|} \exp(\langle\overline{\boldsymbol{x}}_c\rangle^T\langle\boldsymbol{e}'_{c'}\rangle / \tau)},
    \label{equ:glcl}
\end{equation}
where $\langle\boldsymbol{x}\rangle=\boldsymbol{x}/\Vert \boldsymbol{x}\Vert_2$ is the $l_2$-normalization operator, and $\tau$ is the temperature hyper-parameter.

\subsection{Training and Inference}
\label{sec:inf}
\noindent\textbf{- Training.} Since the pseudo labeling always requires a relatively good initialization, we utilize a two-stage training strategy. The first stage only uses the reliable labels, \ie, the video-level labels, to pretrain the network. Hence the loss function is $\mathcal{L}_{\text{I}}=\mathcal{L}_{\text{vid}}$. The loss for the second stage further includes the frame-wise losses on both classification and representation using pseudo labels: $\mathcal{L}_{\text{II}}=\alpha\mathcal{L}_{\text{vid}} + \beta\mathcal{L}_{\text{cls}} + \gamma\mathcal{L}_{\text{glc}}$, where $\alpha$, $\beta$ and $\gamma$ are all hyper-parameters.

\noindent\textbf{- Inference.} For inference, we obtain the action label $\hat{y}_t$ for frame $t$ directly from the frame-wise class probabilities: 
$\hat{y}_t=\arg\max_c \boldsymbol{P}_{t,c}$. Note that we do not require any alignment processing during inference, which is performed by previous WSAS methods for smoother predictions.
However, since the ground-truth transcript is not available during inference, these methods either run alignment with every transcript from the training set~\cite{richard2018neuralnetwork, li2019weakly, chang2019d3tw, lu2021weakly, chang2021learning}, or predict the transcript via another model~\cite{souri2021fast}, which makes the pipeline more time-consuming and less practical. 

\section{Experiments}

\subsection{Experimental Setup}
\label{sec:setup}

\noindent\textbf{- Datasets.} We perform experiments on three datasets. The \textbf{Breakfast}~\cite{kuehne2014language} dataset contains 1712 videos of breakfast cooking with 48 different actions. On average, each video has 6.8 segments and 7.3\% frames are background. The \textbf{Hollywood Extended}~\cite{bojanowski2014weakly} dataset contains 937 videos taken from movies with 16 categories of daily actions such as \textit{walk} or \textit{sit}. On average, each video has 5.9 segments and 60.9\% frames are background. The \textbf{CrossTask}~\cite{zhukov2019cross} dataset contains videos from 18 primary tasks. Following~\cite{lu2021weakly}, 14 cooking-related tasks are selected, which have 2552 videos and 80 action categories. On average, each video has 14.4 segments and 74.8\% frames are background. 
For Breakfast, we use the released 4 training/test splits and report the average. For Hollywood, we perform a 10-fold cross-validation. For CrossTask, we use the released training/test split. These evaluation protocols are consistent with previous methods.

\noindent\textbf{- Metrics.} To evaluate our method, we use 4 standard metrics, following \cite{ding2018weakly, li2019weakly, chang2021learning, lu2021weakly}. (1) The \textit{Mean-over-Frames} (\textbf{MoF}) is the percentage of frames whose labels are correctly predicted. (2) The \textit{Mean-over-Frames without Background} (\textbf{MoF-Bg}) is the MoF over non-background frames. It is more suitable than MoF for the datasets with high background rate such as Hollywood and CrossTask. (3) The \textit{Intersection-over-Union} (\textbf{IoU}) is defined as $|I\cap I^*|/|I\cup I^*|$ while (4) the \textit{Intersection-over-Detection} (\textbf{IoD}) is $|I\cap I^*|/|I|$, where $I^*$ and $I$ are the ground-truth (GT) segment and the predicted segment \textit{with the same class}, respectively. For each GT segment, the highest IoU/IoD with \textit{one} predicted segment are preserved, and the average of all GT segments is reported. Note that the definition of the IoU/IoD in \cite{lu2021weakly} is different from other works \cite{ding2018weakly, li2019weakly, chang2021learning}, and will be indicated with special symbols when used.

\noindent\textbf{- Input Features.} As the input features $\boldsymbol{X}$ for Breakfast and Hollywood, we use the 2048-dimensional RGB+flow I3D features~\cite{carreira2017quo} adopted by MuCon \cite{souri2021fast} and most FSAS methods \cite{farha2019ms, li2020ms, liu2023diffusion, yi2021asformer}, while some WSAS methods \cite{richard2018neuralnetwork, li2019weakly, lu2021weakly, chang2021learning} still use the iDT features~\cite{wang2013action}. Since recent studies~\cite{souri2021fast, souri2020evaluating} have found that they perform worse with more advanced I3D features, we still report the performance with the iDT features for them following~\cite{souri2021fast}. For CrossTask, the officially released 3200-dimensional features are adopted, but we do not perform dimension reduction via PCA as with~\cite{lu2021weakly}. For GPU memory efficiency, we perform a 10$\times$ temporal downsampling for Breakfast and 5$\times$ for Hollywood. During inference, the output is upsampled to match the original video/ground-truth length.

\noindent\textbf{- Implementation Details.} We adopt a 6-layer Transformer \cite{vaswani2017attention} with single-head self-attention as the temporal network, and the latent dimension $d'$ is 256. For the ATBA, we set $w^\text{b}=7$ and $\lambda=4$ for all datasets. $w^\text{a}$ is set to 31 for Breakfast/CrossTask and 23 for Hollywood.  $\mu$ is set to 0.3 for Breakfast/Hollywood and 0.1 for CrossTask. The loss weights $\alpha$, $\beta$ are set to 1.0 and $\gamma$ is 0.1. $\tau$ in \cref{equ:glcl} is set to 0.2. Besides, to alleviate the issue that the predictions on Hollywood/CrossTask are overly dominated by background, we lower the sample weights of \textit{pseudo} background frames to 0.8 in \cref{eq:cls} for these datasets. 

\noindent\textbf{- Multiple Runs.} Due to the alternating nature of learning from weak supervision, there are often fluctuations in the results of WSAS methods~\cite{souri2021fast, lu2021weakly}. Hence, following~\cite{souri2021fast}, we report the average and standard deviation over 5 runs with different random seeds for better evaluation.

\begin{table}
    \centering
    \resizebox{\linewidth}{!}{
    \begin{tabular}{c|cccc}
        \toprule
        \multicolumn{5}{c}{\textbf{Breakfast}}\\
        \midrule
         Method & MoF\textcolor{gray}{\scriptsize $\pm$std} & MoF-Bg\textcolor{gray}{\scriptsize $\pm$std} & IoU\textcolor{gray}{\scriptsize $\pm$std} & IoD\textcolor{gray}{\scriptsize $\pm$std}\\
         \midrule
        HMM+RNN \cite{richard2017weakly} & 33.3 & - & - & - \\
        \cite{richard2017weakly}+Length \cite{kuehne2018hybrid} & 36.7 & - & - & - \\
        TCFPN+ISBA \cite{ding2018weakly} & 38.4\big/36.4\textcolor{gray}{\scriptsize $\pm$1.0}* & 38.4 & 24.2 & 40.6\\
        NN-Viterbi \cite{richard2018neuralnetwork} & 43.0\big/39.7\textcolor{gray}{\scriptsize $\pm$2.4}* & - & - & - \\
        D3TW \cite{chang2019d3tw} & 45.7 & - & - & - \\
        CDFL \cite{li2019weakly} & 50.2\big/48.1\textcolor{gray}{\scriptsize $\pm$2.5}* & 48.0 & 33.7 & 45.4 \\
        DPDTW \cite{chang2021learning} & 50.8 & - & 35.6 & 45.1\\
        TASL \cite{lu2021weakly} $\spadesuit$ & 47.8 & - & 35.2\dag & \underline{46.1}\dag\\
        MuCon \cite{souri2021fast} $\spadesuit$ & 48.5\textcolor{gray}{\scriptsize $\pm$1.8} & \underline{50.3}* & \underline{40.9}* & \underline{54.0}* \\
        POC \cite{lu2022set} $\spadesuit$ & 45.7 & - & \underline{38.3}\dag & -\\
		AdaAct \cite{zhang2023hoi} & \underline{51.2} & 48.3 & 36.3 & 46.4\\
        \midrule
        {\textbf{ATBA} $\spadesuit$} & \multirow{2}{*}{\textbf{53.9}\textcolor{gray}{\scriptsize $\pm$1.2}} & \multirow{2}{*}{\textbf{54.4}\textcolor{gray}{\scriptsize $\pm$1.2}} & \textbf{41.1}\textcolor{gray}{\scriptsize $\pm$0.7} & \textbf{61.7}\textcolor{gray}{\scriptsize $\pm$1.1}\\
        \textbf{(Ours)} & & & \textbf{39.5}\textcolor{gray}{\scriptsize $\pm$0.8}\dag &\textbf{55.9}\textcolor{gray}{\scriptsize $\pm$1.0}\dag\\
        \bottomrule\addlinespace[1pt]
    
        \toprule
        \multicolumn{5}{c}{\textbf{Hollywood Extended}}\\
        \midrule
         Method & MoF\textcolor{gray}{\scriptsize $\pm$std} & MoF-Bg\textcolor{gray}{\scriptsize $\pm$std} & IoU\textcolor{gray}{\scriptsize $\pm$std} & IoD\textcolor{gray}{\scriptsize $\pm$std}\\
         \midrule
        HMM+RNN \cite{richard2017weakly} & - & - & 11.9 & - \\
        \cite{richard2017weakly}+Length \cite{kuehne2018hybrid} & - & - & 12.3 & - \\
        TCFPN+ISBA \cite{ding2018weakly} & 28.7 & 34.5 & 12.6 & 18.3\\
        D3TW \cite{chang2019d3tw} & 33.6 & - & - & - \\
        CDFL \cite{li2019weakly} & 45.0 & \underline{40.6} & 19.5 & 25.8 \\
        DPDTW \cite{chang2021learning} & \textbf{55.6} & 25.6$\sharp$ & \textbf{33.2} & \underline{43.3}\\
        TASL \cite{lu2021weakly} & 42.1$\sharp$ & 27.2$\sharp$ & \underline{23.3}\dag$\sharp$ & \underline{33.0}\dag$\sharp$\\
        MuCon \cite{souri2021fast} $\spadesuit$ & - & \textbf{41.6} & 13.9* & - \\
        \midrule
        {\textbf{ATBA} $\spadesuit$} & \multirow{2}{*}{\underline{47.7}\textcolor{gray}{\scriptsize $\pm$2.5}} & \multirow{2}{*}{{40.2}\textcolor{gray}{\scriptsize $\pm$1.6}} & \underline{30.9}\textcolor{gray}{\scriptsize $\pm$1.6} & \textbf{55.8}\textcolor{gray}{\scriptsize $\pm$0.8}\\
        \textbf{(Ours)} & & & \textbf{28.5}\textcolor{gray}{\scriptsize $\pm$1.6}\dag &\textbf{44.9}\textcolor{gray}{\scriptsize $\pm$0.5}\dag\\
        \bottomrule\addlinespace[1pt]

        \toprule
        \multicolumn{5}{c}{\textbf{CrossTask}}\\
        \midrule
         Method & MoF\textcolor{gray}{\scriptsize $\pm$std} & MoF-Bg\textcolor{gray}{\scriptsize $\pm$std} & IoU\textcolor{gray}{\scriptsize $\pm$std} & IoD\textcolor{gray}{\scriptsize $\pm$std}\\
         \midrule
        NN-Viterbi \cite{richard2018neuralnetwork} $\spadesuit$ & 26.5* & - & 10.7\dag* & 24.0\dag* \\
        CDFL \cite{li2019weakly} $\spadesuit$ & 31.9* & - & 11.5\dag* & 23.8\dag* \\
        TASL \cite{lu2021weakly} $\spadesuit$ & 40.7 & \underline{27.4}$\sharp$ & 14.5\dag & \textbf{25.1}\dag\\
        POC \cite{lu2022set} $\spadesuit$ & \underline{42.8} & 17.6$\sharp$ & \underline{15.6}\dag & - \\
        \midrule
        {\textbf{ATBA} $\spadesuit$} & \multirow{2}{*}{\textbf{50.6}\textcolor{gray}{\scriptsize $\pm$1.3}} & \multirow{2}{*}{\textbf{31.3}\textcolor{gray}{\scriptsize $\pm$0.7}} & \textbf{20.9}\textcolor{gray}{\scriptsize $\pm$0.4} & \textbf{44.6}\textcolor{gray}{\scriptsize $\pm$0.7}\\
        \textbf{(Ours)} & & & \textbf{15.7}\textcolor{gray}{\scriptsize $\pm$0.3}\dag &\underline{24.6}\textcolor{gray}{\scriptsize $\pm$0.4}\dag\\
        \bottomrule
    \end{tabular}}
    
    \caption{Comparisons of ours with other WSAS methods on three datasets. \textcolor{gray}{std} is the standard deviation of multiple runs (if any). \dag-The metric is computed by the definition of \cite{lu2021weakly}. *-Results are reported by other works. Please refer to the supplementary material for detailed sources. $\sharp$-Results are obtained by us via rerunning the open sources (The TASL~\cite{lu2021weakly} does not follow the common 10-fold evaluation protocol for Hollywood so we re-produce the results). $\spadesuit$-The reported results are the average of multiple runs. Best results are in bold, second best are underlined.}
    \label{tab:sota}
\end{table}

\begin{table}
    \centering
    \resizebox{0.9\linewidth}{!}{
    \begin{tabular}{c|c|ccc}
        \toprule
        \multirow{2}{*}{Method} & \multirow{2}{*}{Tr.A.} & \multirow{2}{*}{MoF} & Training  & Inference \\
        & & & (Hours) & (Seconds)\\
        \midrule
        TCFPN+ISBA \cite{ding2018weakly} & \ding{55} & 33.3 & 12.75* & \textbf{0.01}*\\
        NN-Viterbi \cite{richard2018neuralnetwork} & V & 43.0 & 11.23* & 56.25*\\
        CDFL \cite{li2019weakly} & V & 50.2 & 66.73* & 62.37*\\
        DPDTW \cite{chang2021learning} & D & \underline{50.8} & 31.02$\sharp$ & \underline{0.69}$\sharp$\\
        TASL \cite{lu2021weakly} & V & 47.8 & 24.66$\sharp$ & 54.99$\sharp$\\
        MuCon \cite{souri2021fast} & \ding{55} & 48.5 & 4.57 & 3.03\\
        POC \cite{lu2022set} & \ding{55} & 45.7 & \textbf{2.28}$\sharp$ & \textbf{0.01}$\sharp$\\
        \midrule
        \textbf{ATBA (Ours)} & B & \textbf{53.9} & \underline{3.45} & \textbf{0.01}\\
        \bottomrule
    \end{tabular}
    }
    \caption{Comparison of accuracy, training and inference time on the Breakfast. The training time is measured as the entire training duration on the split 1, and the inference time is measured as the average time for inferring a video from the test set of the split 1. Tr.A.-The alignment algorithm adopted during training (\ding{55}-No Alignment. V-Viterbi. D-DTW. B-Boundary Alignment). *-Measured by \cite{souri2021fast}. $\sharp$-Measured by us. Best results are in bold, second best are underlined.}
     \label{tab:eff}
\end{table}

\subsection{Comparison with the State-of-the-Art}
\label{sec:cmp}
\noindent\textbf{- Performance.} In \Cref{tab:sota}, we compare our proposed method with previous WSAS methods. Our ATBA achieves state-of-the-art (SOTA) by a clear margin (+2.7\% MoF) on the Breakfast~\cite{kuehne2014language}, demonstrating the effectiveness of focusing on action transitions. Moreover, comparing the standard deviation with other methods, it can be found that our method is also relatively stable.

Our ATBA also achieves comparable performance on the Hollywood~\cite{bojanowski2014weakly}. The reason why our method does not show significant advantage is probably because this dataset is collected from movies and so contains many shot changes, resulting in more noisy boundaries. Note that although the DPDTW~\cite{chang2021learning} achieves significantly high MoF, this metric is severely biased in case of highly imbalanced categories (60.9\% frames are background). We report the MoF-Bg metric for it, 
on which it performs poorly, proving that it cannot recognize real actions very well. In contrast, the performance of our method is more balanced.

For the more difficult dataset CrossTask~\cite{zhukov2019cross} with the most segments and highest background rate, our approach also outperforms previous methods whether or not the metric involves background (+7.8\% MoF and +3.9\% MoF-Bg).

\noindent\textbf{- Efficiency.} Besides of the performance, we also compare the training time of our approach with previous WSAS methods to show the efficiency of ours. Following~\cite{souri2021fast}, we train our model on an Nvidia GeForce GTX 1080Ti GPU, and the training time is measured as the wall time over the whole training phase, during which any irrelevant operations such as intermediate evaluation and saving checkpoints are removed. As all the WSAS methods directly load pre-computed features, the time measurement also does not include the time to extract features from raw videos. As shown in \Cref{tab:eff}, our ATBA achieves the best performance with the second shortest training time, which demonstrates the effectiveness of our design. Specifically, the training speed of ours is on average 10 times (varying from 3 to 20) faster than methods performing frame-by-frame alignment \cite{richard2018neuralnetwork, li2019weakly, chang2021learning, lu2021weakly}. Compared to the alignment-free methods with comparable training speed \cite{souri2021fast, lu2022set}, ours achieves better performance.

We also compare the time to infer a single test video. As mentioned in \cref{sec:inf}, ours does not require any alignment processing, which makes it the fastest during inference.

\subsection{Ablation Studies}
\label{sec:aba}
\noindent\textbf{- Effect of ATBA.} In \Cref{tab:atba}, we conducted an in-depth evaluation of our proposed ATBA. From Exp.1, the pseudo label quality and the evaluation performance are both very poor when only using the class-agnostic boundary scores $\mathcal{V}^\text{b}$ to directly select $M-1$ boundaries for transitions via the greedy strategy stated in \cref{sec:atba}, showing that it is critical to take the class-specific transition pattern into account as with our design to suppress the noisy boundaries for more precise pseudo segmentation. \cref{fig:intro_bdy} provides an intuitive example. In addition, comparing Exp.2 and 3, it's better to involve the class-agnostic boundary scores in the action transition alignment. We think it is because that some candidates can have very low boundary scores (unlikely to be a boundary) as the candidate selection only depends on ranking, and when the transition patterns are not discriminative at the beginning of training, these candidates may be unexpectedly aligned if the boundary scores are not involved.

\noindent\textbf{- Effect of Video-Level Losses.} In \Cref{tab:g2l}, we evaluate the $\mathcal{L}_\text{vid}$ and $\mathcal{L}_\text{glc}$. Comparing Exp.1 and 2, the model with $\mathcal{L}_\text{vid}$ achieves higher performance (+6.5\% MoF) at the same level of pseudo label accuracy, demonstrating that the video-level supervision can promote the learning of precise action semantics. In Exp.3, the explicit representation alignment ($\mathcal{L}_\text{glc}$) further improves the quality of frame features and thus the performance (+1.6\% MoF). 
Moreover, Exp.4 shows that whether or not to share the classifier between frame-/video-level classification has little impact. 

\begin{table}
    \centering
    \resizebox{0.8\linewidth}{!}{
    \begin{tabular}{c|c|cc|c|ccc}
        \toprule
        \multirow{2}{*}{Exp.} & \multirow{2}{*}{$\mathcal{V}^\text{b}$} & \multicolumn{2}{c|}{Cost Mat.} & \multirow{2}{*}{P.L.} & \multirow{2}{*}{MoF} & \multirow{2}{*}{IoU} & \multirow{2}{*}{IoD}\\
        & & $\boldsymbol{V}^\text{a}$ & $\boldsymbol{V}$ & & & &\\
        \midrule
        1 & \checkmark & & & 38.6 & 32.3 & 23.1 & 52.5\\
        2 & \checkmark & \checkmark & & 51.7& 41.3 & 30.2 & 47.8\\
        \midrule
        \rowcolor{bgblue} 3 & \checkmark & $\circ$ & \checkmark & \textbf{67.9} & \textbf{54.0} & \textbf{41.1} & \textbf{62.3}\\
        \bottomrule
    \end{tabular}
    }
    \caption{Ablation studies of ATBA on the Breakfast. Exp.-Different experiment configurations. Cost Mat.-Different choices for the cost matrix in action transition alignment. $\circ$-According to \cref{eq:add}, $\boldsymbol{V}$ contains $\boldsymbol{V}^\text{a}$. P.L.-The accuracy of pseudo labels during training. Exp.1 means that only the class-agnostic boundary detection is adopted to localize action transitions. \sethlcolor{bgblue}\hl{Exp.3 is our default setting.} Best results are in bold.}
     \label{tab:atba}
\end{table}

\begin{table}
    \centering
    \resizebox{0.85\linewidth}{!}{
    \begin{tabular}{c|cc|c|c|ccc}
        \toprule
        Exp. & $\mathcal{L}_\text{vid}$ & $\mathcal{L}_\text{glc}$ & D.Cls.& P.L. & MoF & IoU & IoD\\
        \midrule
        1 & & & & 67.5 & 45.9 & 40.5 & 61.9\\
        2 & \checkmark & & & 66.6 & 52.4 & 38.8 & 61.4\\
        \midrule
        \rowcolor{bgblue} 3 & \checkmark & \checkmark & & 67.9 & \textbf{54.0} & \textbf{41.1} & \textbf{62.3}\\
        \midrule
        4 & \checkmark & \checkmark & \checkmark & \textbf{68.1} & 53.7 & 40.1 & 61.7\\
        \bottomrule
    \end{tabular}
    }
    \caption{Ablation studies of video-level losses on the Breakfast. Exp.-Different experiment configurations. D.Cls.-Different classifiers for frame-wise prediction and action occurrence prediction. P.L.-The accuracy of pseudo labels during training. \sethlcolor{bgblue}\hl{Exp.3 is our default setting.} Best results are in bold.}
     \label{tab:g2l}
\end{table}

We also provide the analysis on the effect of important hyper-parameters in the supplementary material, including $w^\text{b}$, $w^\text{a}$, $\mu$ and $\lambda$ in the ATBA.

\begin{figure}
    \centering
    \includegraphics[width=\linewidth]{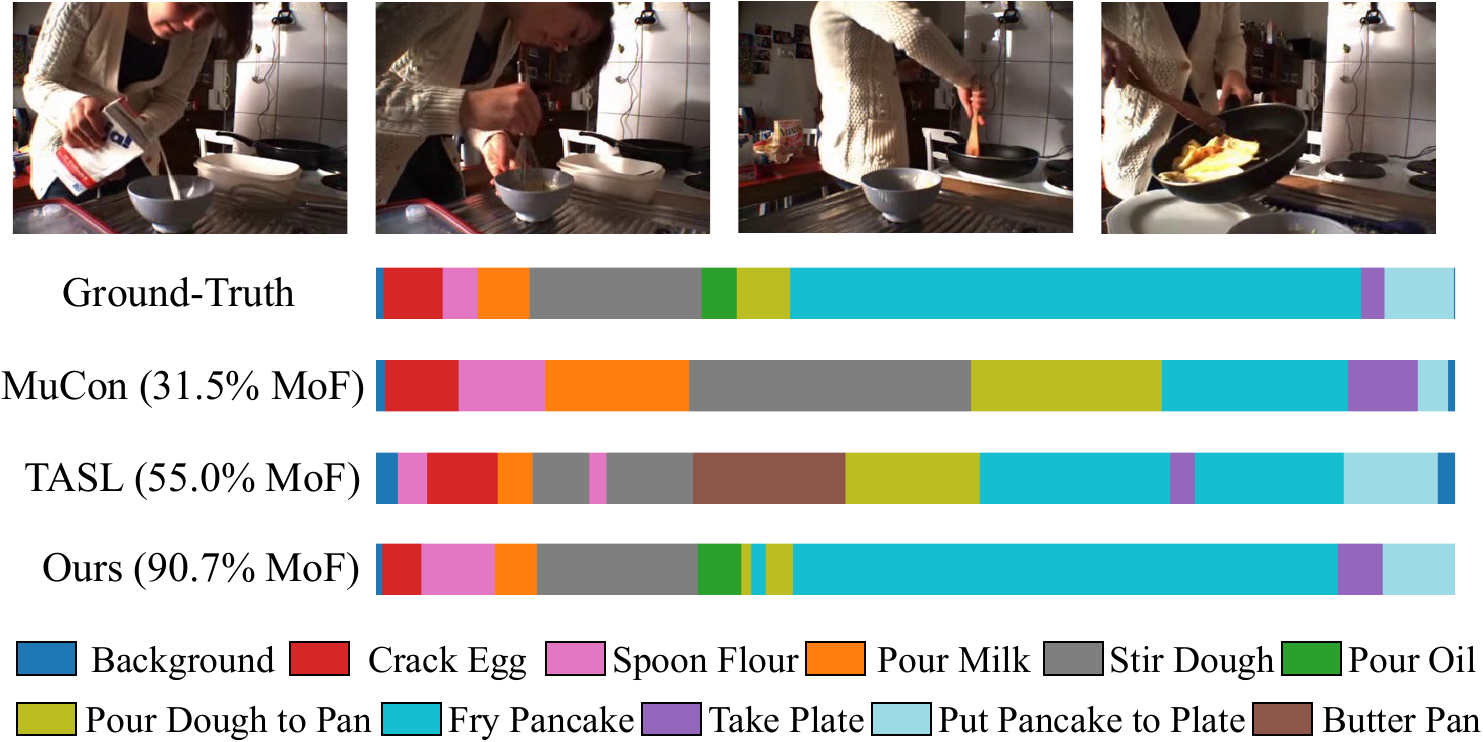}
    \caption{Qualitative results on the Breakfast. The example test video is \textit{P15-stereo01-P15-pancake}. We compare our inference result with two recent methods. Best viewed in color.}
    \label{fig:test}
\end{figure}

\subsection{Qualitative Results}
\label{sec:vis}

We show the qualitative inference result compared with two recent open source WSAS methods, \ie, MuCon \cite{souri2021fast} and TASL \cite{lu2021weakly}, in~\cref{fig:test}. Our ATBA achieves significantly more accurate result in this challenging video with many segments. The result of TASL shows order reversal (``\textit{Crake Egg}" and ``\textit{Spoon Flour}") and hallucination (``\textit{Butter Pan}"). The MuCon predicts the correct action ordering, but the result deviates severely. In contrast, ours successfully predicts an accurate segmentation, indicating the action semantics are well learned. More qualitative results are provided in supplementary material.

\section{Conclusion}
In this work, we propose to directly localize action transitions for efficient pseudo segmentation generation in the WSAS task, thus avoiding the time-consuming frame-by-frame alignment. Due to the presence of noisy boundaries, a novel Action-Transition-Aware Boundary Alignment (ATBA) framework is proposed to efficiently and effectively filter out noise and detect transitions. Moreover, we also design some video-level losses to utilize video-level supervision to improve the semantic robustness. Extensive experiments show the effectiveness of our ATBA.

\section{Acknowledgment}
This work was supported partially by the NSFC (U21A20471), National Key Research and Development Program of China (2023YFA1008503), Guangdong NSF Project (No. 2023B1515040025, 2020B1515120085).

{
    \small
    \bibliographystyle{ieeenat_fullname}
    \bibliography{main}
}

\clearpage
\maketitlesupplementary

\section{Details of Temporal Network}
As the recent works~\cite{yi2021asformer, du2022efficient} have pointed out, the vanilla self-attention mechanism is not suitable for action segmentation task, since it is hard to be learned to focus on meaningful temporal positions over a very long video. Hence, we replace the vanilla self-attention with a pyramid hierarchical local attention as in \cite{yi2021asformer} to achieve a local-to-global learning pattern which is similar to CNNs. Specifically, each frame only performs self-attention with the frames in a local window centered at itself, and the window size increases in the deeper layers. The radius of the window is set to $2^{l-1}$ in the $l$-th (beginning from 1) encoder layer.

\section{Construction of Pairwise Similarity Matrix}
In the class-agnostic boundary scoring step of our Action-Transition-Aware Boundary Alignment (ATBA), a pairwise similarity matrix $\boldsymbol{\Gamma}^{(t)} \in \mathbb{R}^{w^\text{b}\times w^\text{b}}$ is calculated within the local window with size $w^\text{b}$ centered at $t$, from the model output $\boldsymbol{P}$:
\begin{equation}
\begin{gathered}
    \boldsymbol{\Gamma}^{(t)}_{i,j} = 1-2\ \text{JS}(\boldsymbol{p}_{\text{ind}^\text{b}(t,i)}, \boldsymbol{p}_{\text{ind}^\text{b}(t,j)}),\ 1\leq i,j \leq w^\text{b},\\
    \text{ind}^\text{b}(t,i)=t-\lfloor\dfrac{w^\text{b}}{2}\rfloor+i-1,
\end{gathered}
\end{equation}
where $\text{ind}^\text{b}(t,i)$ is the index transform from the index $i$ of the local window centered at $t$ to the global timestamp index, $\text{JS}(\cdot,\cdot)$ is the Jensen–Shannon divergence and $\boldsymbol{p}_{\text{ind}^\text{b}(t,i)}$ is the class probability distribution of the frame at timestamp $\text{ind}^\text{b}(t,i)$. $\boldsymbol{\Gamma}^{(t)}_{i,j}$ represents the output similarity between $\text{ind}^\text{b}(t,i)$-th and $\text{ind}^\text{b}(t,j)$-th frames, of which the range is $[-1, 1]$. 

\begin{algorithm}
\caption{Action Transition Alignment}\label{alg:align}
\small
\KwInput{Candidate boundary set $\widetilde{\mathcal{B}}=\{b_k\}_{k=1}^K$; Cost matrix $\boldsymbol{\Delta} \in \mathbb{R}^{K \times (2(M-1)+1)}$}
\Comment{Initialize the cumulative cost matrix $\boldsymbol{D}$}
$\boldsymbol{D} \gets \text{RandomMatrix} \in \mathbb{R}^{K \times (2(M-1)+1)}$\;
\Comment{Initialize the 1st column}
\For{$i\gets 1$ \KwTo $K$}{
\If{$i \leq K-(M-1)$}{
$\boldsymbol{D}_{i,1} \gets \boldsymbol{\Delta}_{i,1}$\;
}
\Else{
$\boldsymbol{D}_{i,1} \gets \infty$\;
}
}

\Comment{Initialize the 2nd column}
\For{$i\gets 1$ \KwTo $K$}{
\If{$i \leq K-(M-1)+1$}{
$\boldsymbol{D}_{i,2} \gets \boldsymbol{\Delta}_{i,2}$\;
}
\Else{
$\boldsymbol{D}_{i,2} \gets \infty$\;
}
}

\Comment{Initialize the 1st row}
\For{$j\gets 3$ \KwTo $2(M-1)+1$}{
$\boldsymbol{D}_{1,j} \gets \infty$\;
}

\Comment{Dynamic programming}
\For{$i\gets 2$ \KwTo $K$}{
\For{$j\gets 3$ \KwTo $2(M-1)+1$}{
\If{$j$ is odd}{
$\boldsymbol{D}_{i,j} \gets \boldsymbol{\Delta}_{i,j}+\min(\boldsymbol{D}_{i-1,j}, \boldsymbol{D}_{i-1,j-1})$\;
}
\Else{
$\boldsymbol{D}_{i,j} \gets \boldsymbol{\Delta}_{i,j}+\min(\boldsymbol{D}_{i-1,j-1}, \boldsymbol{D}_{i-1,j-2})$\;
}
}
}

\Comment{Backtracking}
Initialize the optimal boundary set $\mathcal{B}=\phi$\;
$j \gets 2(M-1) + 1$\;
\For{$i \gets K$ \KwTo $1$}{
\If{$j$ is odd}{
$new\_j \gets \arg\min_{\{j, j-1\}}(\boldsymbol{D}_{i,j}, \boldsymbol{D}_{i,j-1})$\;
}
\Else{
$new\_j \gets \arg\min_{\{j-1, j-2\}}(\boldsymbol{D}_{i,j-1}, \boldsymbol{D}_{i,j-2})$\;
}
\If{$new\_j$ is even}{
Add $b_i$ into $\mathcal{B}$\;
}
$j \gets new\_j$\;
}
Reverse $\mathcal{B}$\;

\KwOutput{Optimal boundary set $\mathcal{B}$}

\end{algorithm}

\section{Details of Action Transition Alignment}
To help better understand the action transition alignment algorithm in our ATBA, we provide the pseudo code in \cref{alg:align}. The algorithm consists of three stages, \ie, initialization (Line 1-20), calculation by dynamic programming (Line 21-30), and backtracking (Line 31-45). The middle stage is stated in the main paper.

\noindent\textbf{- Initialization.} The first row and the first two columns of the cumulative cost matrix $\boldsymbol{D}$ can not be calculated via the recursive equation, and need to be directly initialized before the computation. The rules for the initialization are following:
\begin{itemize}
    \item For the first column, the path through $(k,1)$ means the $k$-th candidate is matched with the \textit{first} empty symbol, \ie, the first $k$ candidates are \textit{all} dropped. However, there are only $K-M+1$ candidates can be dropped, so a valid path cannot pass through the last $M-1$ positions of the first column, so their values are set to $\infty$ (Line 2-9).
    \item The situation in the second column is similar to the first column, where a path through $(k, 2)$ means that the $k$-th candidate is matched with the first transition. To ensure that the remaining $M-2$ transitions can be matched, at least the last $M-2$ candidates cannot be matched with the first transition. Hence the values of the last $M-2$ entries of the second column are set to $\infty$ (Line 10-17).
    \item For the first row, as mentioned in the main paper, only $(1,1)$ and $(1,2)$ are valid, so the values of other entries in the first row are set to $\infty$ (Line 18-20).
    \item The remaining valid positions can be initialized with the values of corresponding entries in $\boldsymbol{\Delta}$, as these positions are all in the first two columns, and so relevant to at most one transition matching without accumulating multiple costs.
\end{itemize}


\noindent\textbf{- Backtracking.} After filling the matrix $\boldsymbol{D}$, we find out the optimal boundary set $\mathcal{B}$ (\ie, an alignment path) from it using backtracking. Clearly, any valid path has exactly one point in each row, meaning that each candidate is matched with one symbol (transition or $\phi$). As mentioned in the main paper, the end position of the optimal path is one of $(K,2(M-1))$ and $(K,2(M-1)+1)$ depending on whose $\boldsymbol{D}$ value is minimal (Line 35 in the first loop, \ie, $i=K$). The backtracking starts from this end position, and runs from bottom to top until the first row. The Line 34-39 calculate the column position in current row $i$ based on the determined position in the next row (\ie, next point in the path). Similar to the forward process, if the column position $j$ of the next row $i+1$ is odd, \ie, the candidate $b_{i+1}$ is dropped, then in current row $i$, candidate $b_i$ can be either dropped ($new\_j=j$) or matched with the previous transition ($new\_j=j-1$) depending on the cumulative cost (Line 34-36). The meaning of Line 37-39 ($j$ is even) is similar. If the point of current row is matched with a transition, we add it into $\mathcal{B}$ (Line 40-42). 

\section{Additional Training Details}
During training, the batch size is 32 and the AdamW \cite{loshchilov2017decoupled} optimizer is adopted. We train the model for 400/300/300 epochs for Breakfast~\cite{kuehne2014language}, Hollywood~\cite{bojanowski2014weakly} and CrossTask~\cite{zhukov2019cross}, respectively, of which the first 40 epochs are the first stage. The initial learning rate is set to 5e-4. The cosine annealing strategy \cite{loshchilov2016sgdr} is used only for the second stage to lower the learning rate to 1/100 of the initial value finally, while the warmup strategy is used for the first 10 epochs of both two stages, beginning from 1/100 and 1/10 of the initial learning rate, respectively. 

\section{Detailed Sources of Results}
In Table 1 of the main paper, some results are not reported by the original paper, and the detailed sources are as follows:

\noindent\textbf{- Breakfast.} The MoF results with standard deviation of ISBA \cite{ding2018weakly}, NN-Viterbi \cite{richard2018neuralnetwork} and CDFL \cite{li2019weakly} are from \cite{souri2021fast}. The MoF-Bg, IoU and IoD results of MuCon \cite{souri2021fast} are from \cite{souri2022fifa}.

\noindent\textbf{- Holloywood Extended.} The IoU result of MuCon \cite{souri2021fast} are from \cite{souri2022fifa}.

\noindent\textbf{- CrossTask.} All the results of NN-Viterbi \cite{richard2018neuralnetwork} and CDFL \cite{li2019weakly} are from \cite{lu2021weakly}.

\begin{figure}
    \centering
    \begin{subfigure}{0.45\linewidth}
    \includegraphics[width=\linewidth]{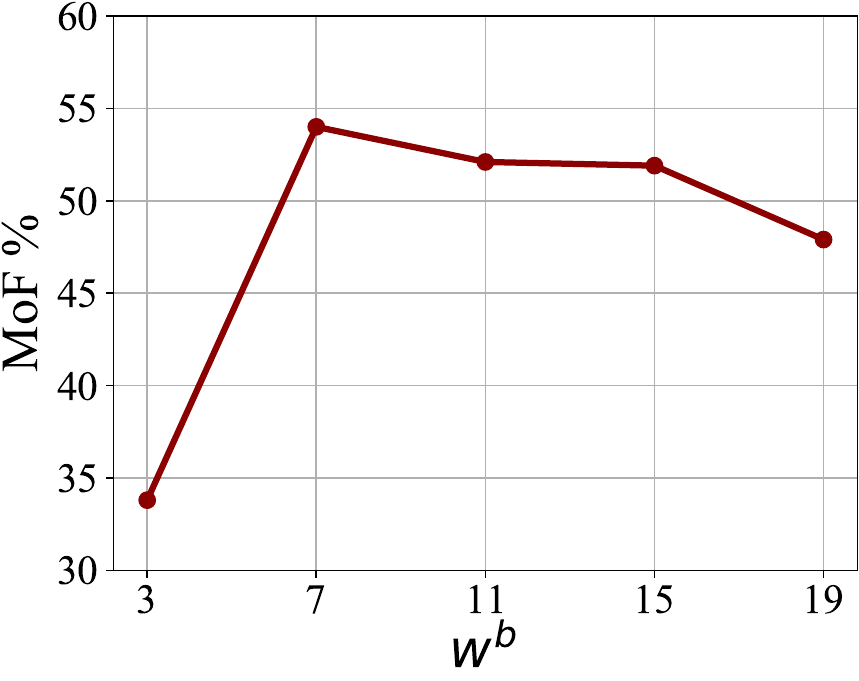}
    \caption{}
    \label{fig:bdyw}
    \end{subfigure}
    \quad
    \begin{subfigure}{0.45\linewidth}
    \includegraphics[width=\linewidth]{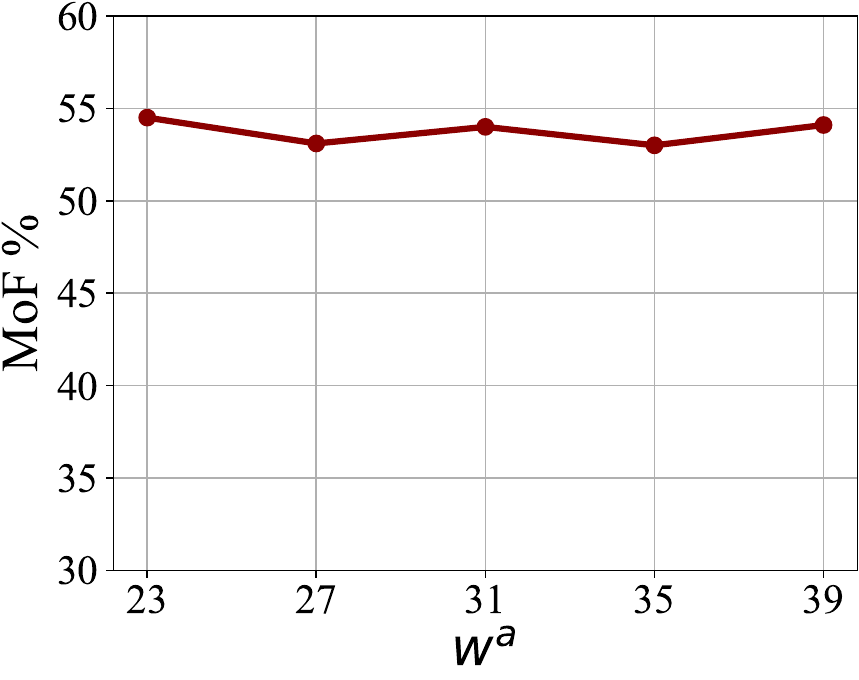}
    \caption{}
    \label{fig:csw}
    \end{subfigure}\\
    \begin{subfigure}{0.45\linewidth}
    \includegraphics[width=\linewidth]{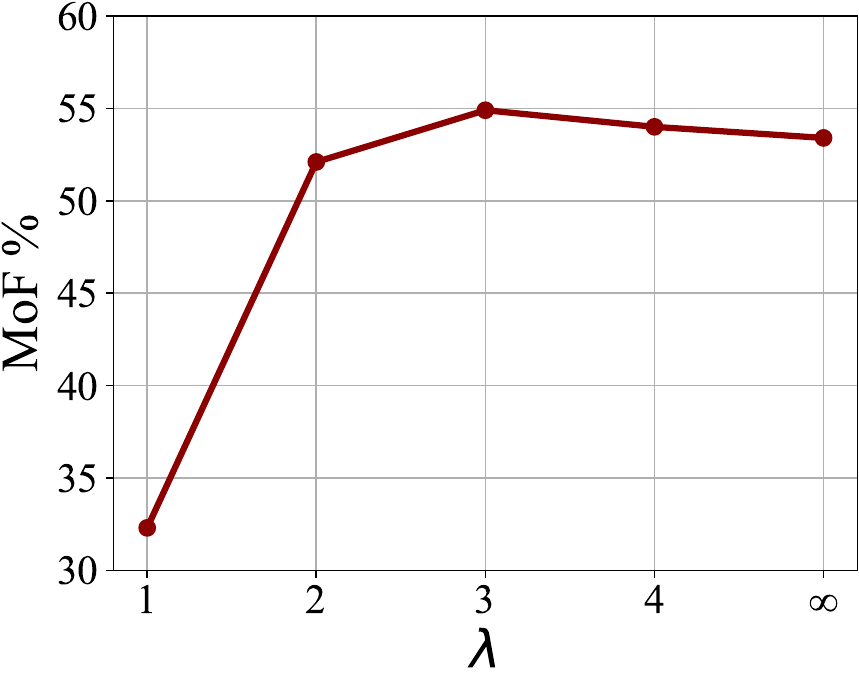}
    \caption{}
    \label{fig:lambda}
    \end{subfigure}
    \quad
    \begin{subfigure}{0.45\linewidth}
    \includegraphics[width=\linewidth]{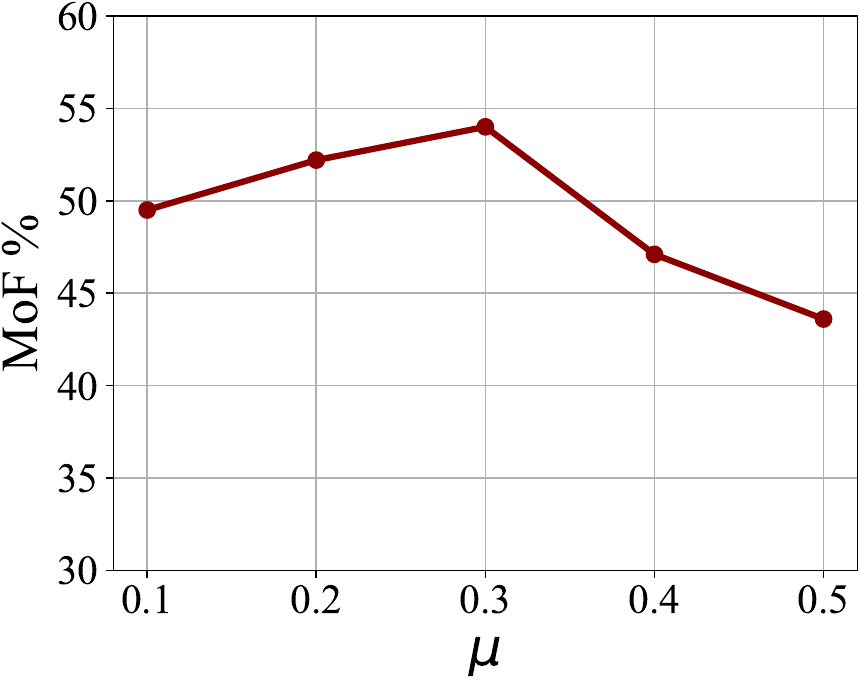}
    \caption{}
    \label{fig:mu}
    \end{subfigure}
    \caption{The Effect of (a) the size of class-agnostic boundary pattern template $w^\text{b}$, (b) the size of action-transition pattern template $w^\text{a}$, (c) $\lambda$, which controls the upper bound of the number of candidate boundaries, and (d) $\mu$, which controls the size of non maximum suppression (NMS) area. The case of $\lambda=\infty$ means that the candidate selection process terminates only when all remaining timestamps are invalid. Experiments are all conducted on the Breakfast.}
    \label{fig:ww}
\end{figure}

\section{Analysis of Hyper-parameters}
\noindent\textbf{- Effect of $w^\text{b}$.} \cref{fig:ww}(a) shows the effect of the size of class-agnostic boundary pattern template $w^\text{b}$. The model performs bad with too small $w^\text{b}$, possibly because it is more susceptible to noise interference. On the other hand, the large $w^\text{b}$ can also lead to performance decrease due to the poor ability of capturing local changes.

\noindent\textbf{- Effect of $w^\text{a}$.} The effect of the size of action transition pattern template $w^\text{a}$ is also shown in \cref{fig:ww}(b). Our method is insensitive to it over a wide range (at least 23-29). Note that these feasible values are much higher than that of $w^\text{b}$, since the action transition scoring aims to capture two adjacent segments which both lasts for a period of time.

\noindent\textbf{- Effect of $\lambda$.} We investigate the effect of $\lambda$ in \cref{fig:ww}(c), which controls the upper bound of the number of candidate boundaries. Note that $\lambda=1$ is equivalent to not applying action transition alignment (\ie, Exp.1 of the ablation study on ATBA in the main paper), so the performance is poor. When $\lambda > 1$, the performance can be maintained at a high level and keep stable as the number of candidates increases, since the additional candidates may be \textit{unambiguous} non-boundary points and have little effect.

\noindent\textbf{- Effect of $\mu$.} \cref{fig:ww}(d) shows the effect of $\mu$, which controls the size of non maximum suppression (NMS) area. Our ATBA prefers relatively small NMS area, since the large NMS area will lead to missing the transitions involving short segments.

\begin{figure}
    \centering
    \begin{subfigure}{\linewidth}
        \includegraphics[width=\linewidth]{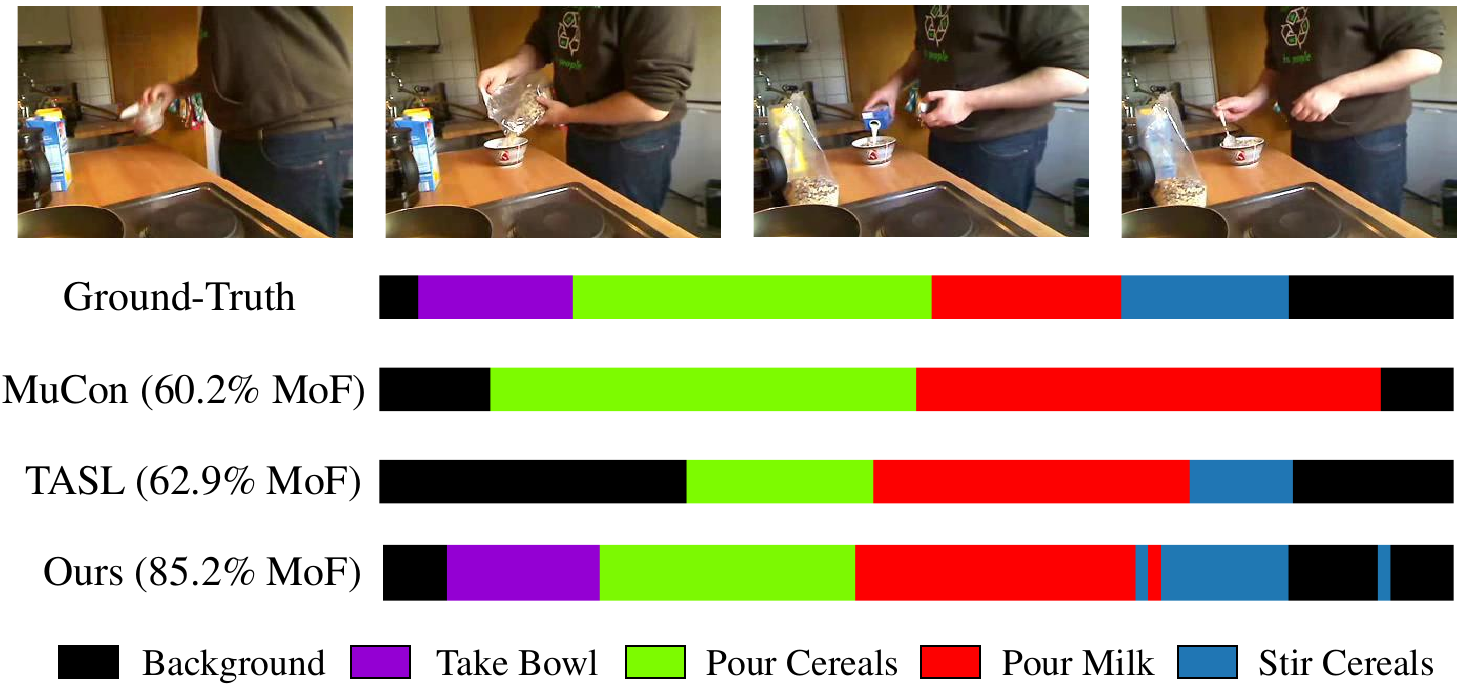}
        \caption{\textit{P03-webcam02-P03-cereals} on Breakfast.}
    \end{subfigure}\\
    \begin{subfigure}{\linewidth}
        \includegraphics[width=\linewidth]{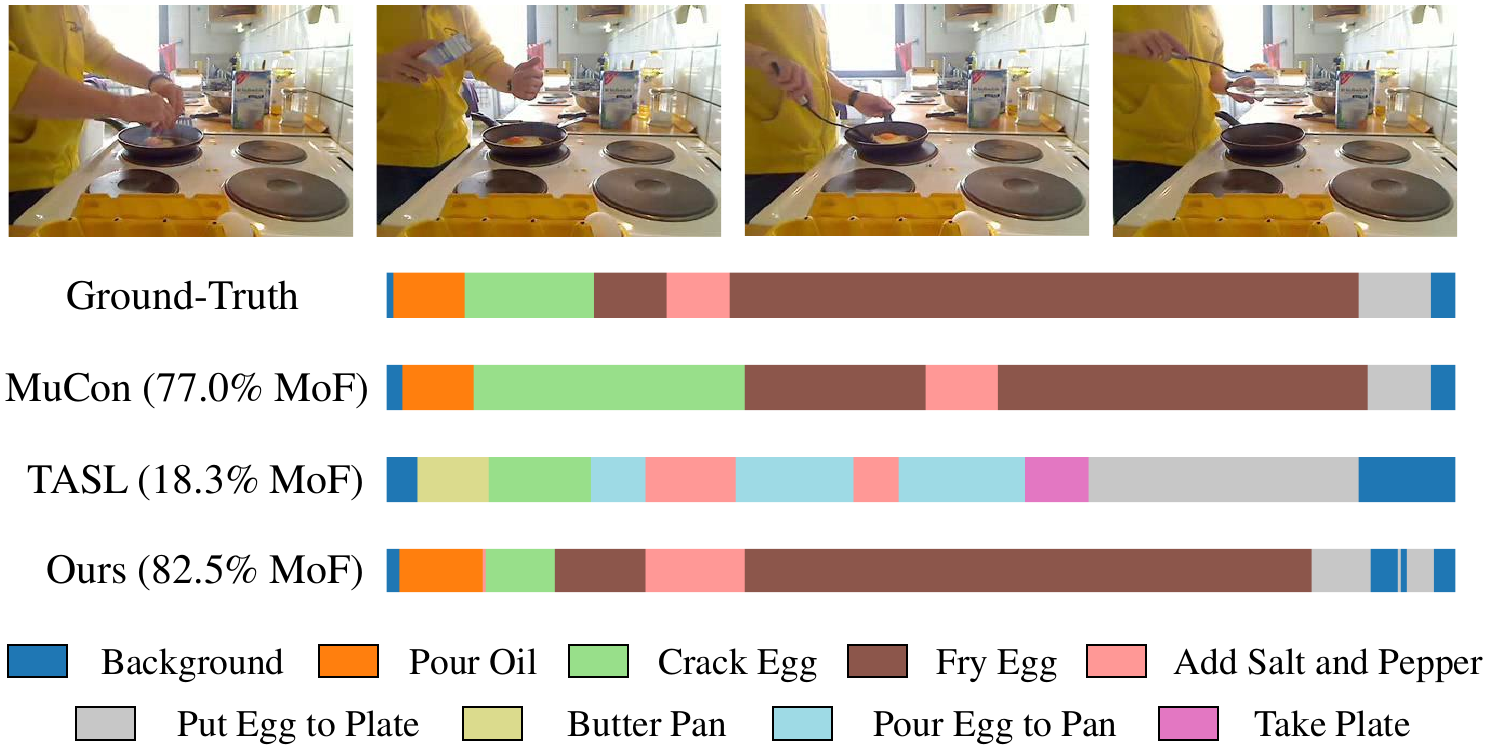}
        \caption{\textit{P10-webcam01-P10-friedegg} on Breakfast.}
    \end{subfigure}\\
    \begin{subfigure}{\linewidth}
        \includegraphics[width=\linewidth]{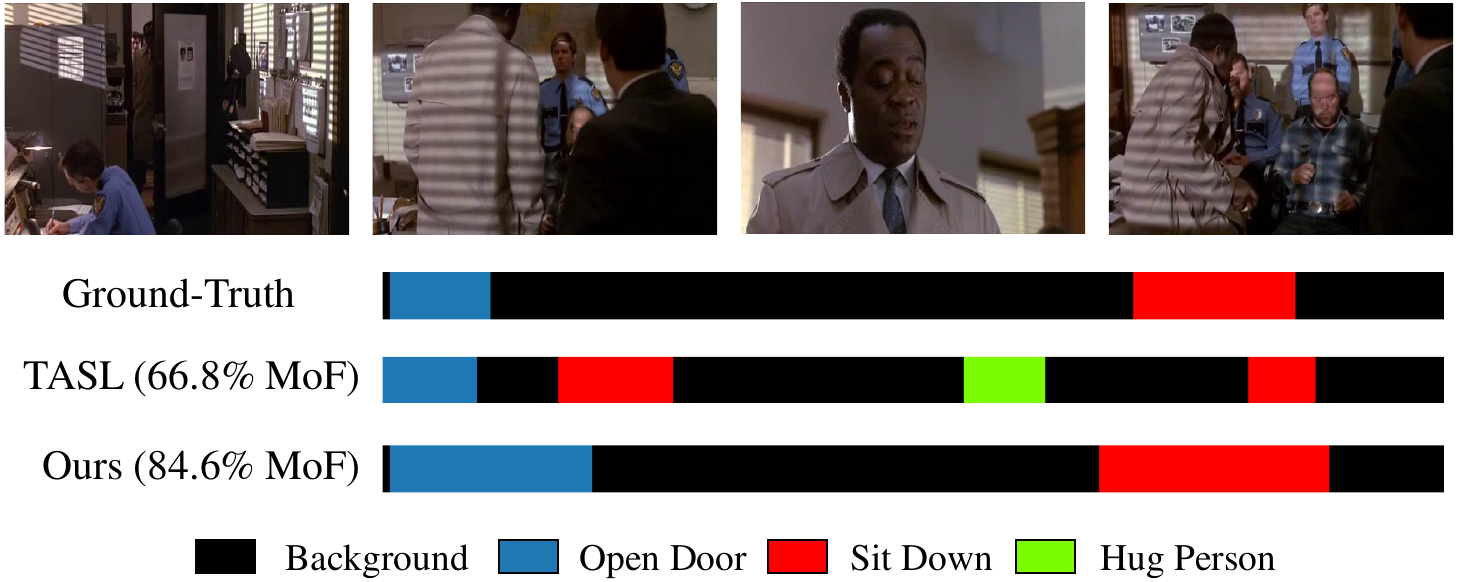}
        \caption{\textit{0896} on Hollywood.}
    \end{subfigure}\\
    \begin{subfigure}{\linewidth}
        \includegraphics[width=\linewidth]{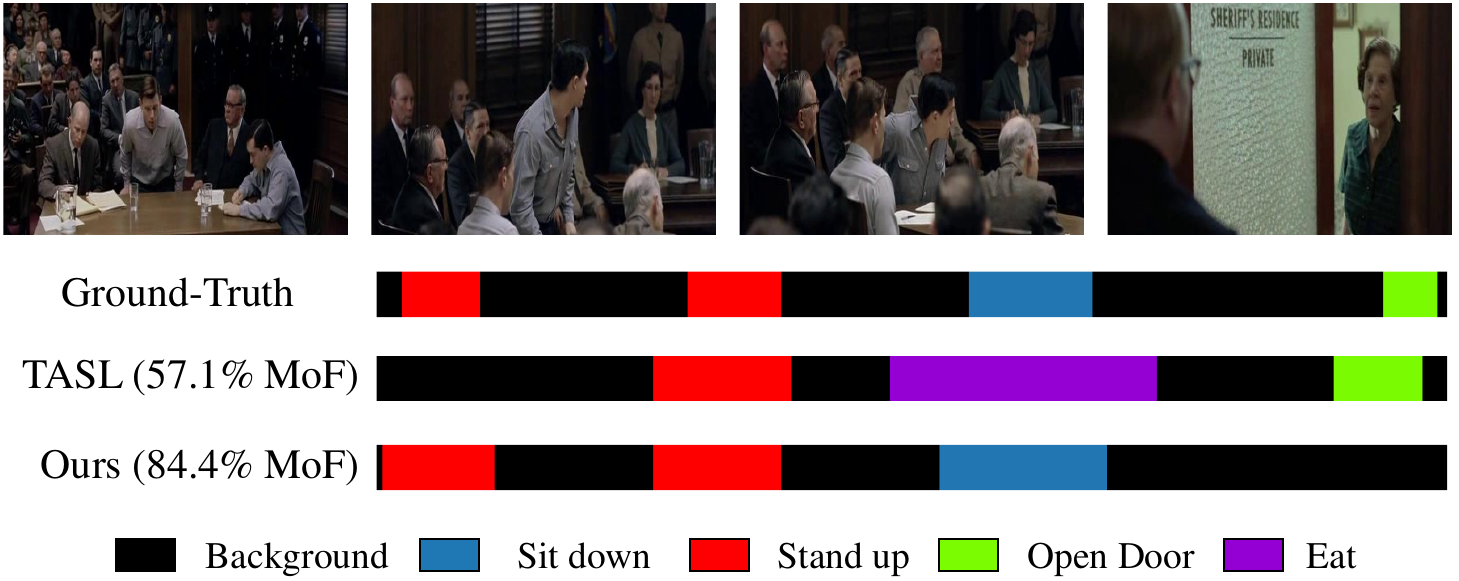}
        \caption{\textit{0146} on Hollywood.}
    \end{subfigure}\\
    \begin{subfigure}{\linewidth}
        \includegraphics[width=\linewidth]{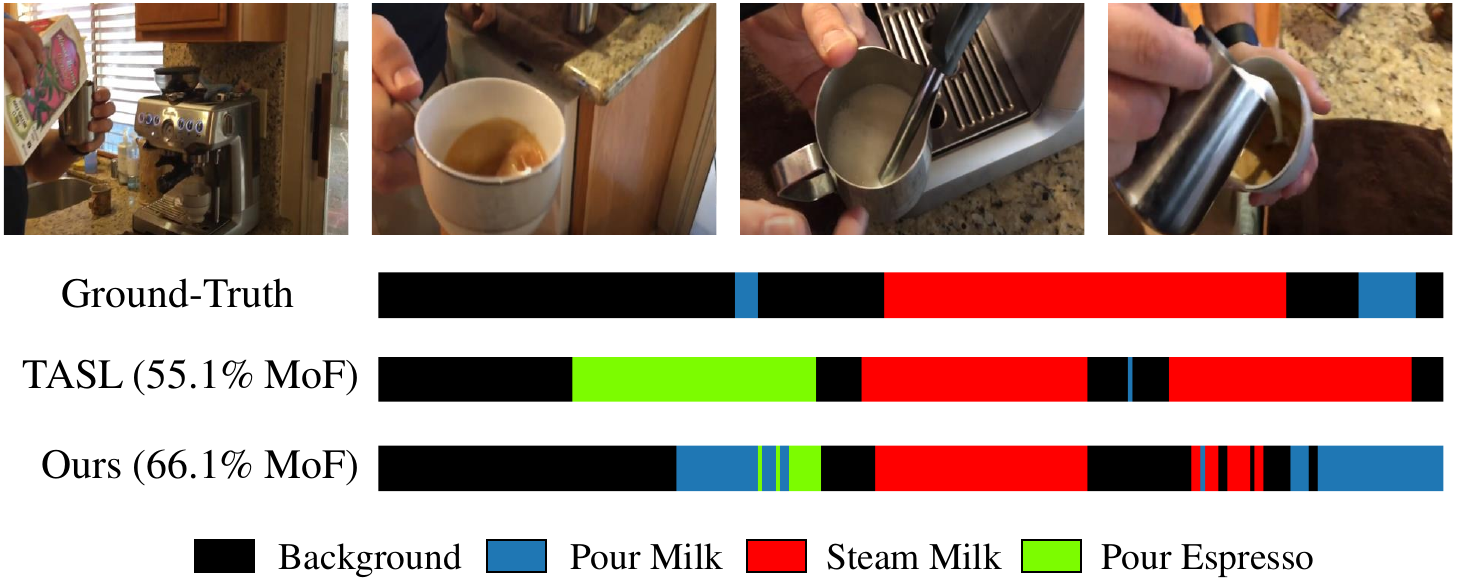}
        \caption{\textit{Make-a-Latte.2qhBLFc5CqM} on CrossTask.}
        \label{fig:test6}
    \end{subfigure}
    \caption{More qualitative results. The names of example test videos are shown below each visualization. Best viewed in color.}
    \label{fig:visual}
\end{figure}

\section{More Qualitative Results}
To help more intuitively understand the advantage of our method, we provide more qualitative results on three datasets: Breakfast \cite{kuehne2014language}, Hollywood \cite{bojanowski2014weakly} and CrossTask~\cite{zhukov2019cross} in \cref{fig:visual}. Our method is significantly more accurate in locating actions than MuCon \cite{souri2021fast} and TASL \cite{lu2021weakly}. Note that in \cref{fig:visual}(e), there is indeed a shot of \textit{espresso} in the video (the 2nd picture, but without a \textit{pouring} action) after action ``\textit{Pour Milk}" (the 1st picture), so the activation on action ``\textit{Pour Espresso}" in our result is not exactly a hallucination compared to the result of TASL \cite{lu2021weakly}.

\end{document}